\documentclass{article}

\PassOptionsToPackage{square, numbers}{natbib}

\usepackage[final]{neurips_2022_ml4ad}

\usepackage[utf8]{inputenc} 
\usepackage[T1]{fontenc}    
\usepackage{hyperref}       
\usepackage{url}            
\usepackage{booktabs}       
\usepackage{amsfonts}       
\usepackage{nicefrac}       
\usepackage{microtype}      
\usepackage{xcolor}         
\usepackage[pdftex]{graphicx}
\usepackage{multirow}
\usepackage{caption}
\usepackage{subcaption}

\title{Rationale-aware Autonomous Driving Policy utilizing Safety Force Field implemented on CARLA Simulator}

\author{%
  Ho Suk\thanks{These authors contributed equally to this work.} \quad Taewoo Kim\footnotemark[1] \quad Hyungbin Park \quad Pamul Yadav \quad Junyong Lee \quad Shiho Kim \\
  Yonsei University \\
  \texttt{\{sukho93, boratw, phb88, pamul, jjunilee, shiho\}@yonsei.ac.kr} \\
}

\begin{document}

\maketitle

\begin{abstract}
Despite the rapid improvement of autonomous driving technology in recent years, automotive manufacturers must resolve liability issues to commercialize autonomous passenger car of SAE J3016 Level 3 or higher. To cope with the product liability law, manufacturers develop autonomous driving systems in compliance with international standards for safety such as ISO 26262 and ISO 21448. Concerning the safety of the intended functionality (SOTIF) requirement in ISO 26262, the driving policy recommends providing an explicit rational basis for maneuver decisions. In this case, mathematical models such as Safety Force Field (SFF) and Responsibility-Sensitive Safety (RSS) which have interpretability on decision, may be suitable. In this work, we implement SFF from scratch to substitute the undisclosed NVIDIA's source code and integrate it with CARLA open-source simulator. Using SFF and CARLA, we present a predictor for claimed sets of vehicles, and based on the predictor, propose an integrated driving policy that consistently operates regardless of safety conditions it encounters while passing through dynamic traffic. The policy does not have a separate plan for each condition, but using safety potential, it aims human-like driving blended in with traffic flow.
\end{abstract}

\section{Introduction}
\label{Section_1}
Guaranteeing the safety of self-driving is one of the essential factors in the autonomous driving system. According to the international standard, ISO 21448, titled Road vehicles – Safety of the intended functionality, the autonomous driving system can be divided into three subsystems: perception, planning, and actuation. As an agent which takes care of safe driving, the driving policy that implements the vehicle level safety strategy (VLSS) at the decision-making level, and the planning subsystem containing the policy are essential. To complete the core of the planning subsystem, developers design and implement driving policies reflecting their own strategy. In 2017, Intel Mobileye introduced Responsibility-Sensitive Safety (RSS) \cite{Shalev-Shwartz 2017}, and NVIDIA introduced Safety Force Field (SFF) \cite{Nister 2017} respectively. Those mathematical models guarantee to avoid going into unsafe situation, ultimately assuring safe driving. The models have two main advantages. First, serving as add-ons, they are compatible with existing driving policy. Second, they are mathematically proven, as a result, actions for safety of vehicle agent are explainable to human robustly.

From the 2010s, ADAS of SAE J3016 Level 2 began to spread widely in the passenger car market, and in 2018, Waymo launched a self-driving taxi service. However, automotive manufacturers in the private car market are struggling with liability issues of autonomous vehicles to achieve Level 3 or higher. It is because a decision failure of autonomous driving system higher than Level 3 could be a responsibility taken by not the driver but the automotive manufacturer who failed to detail the thorough specification for its autonomous driving system. Therefore, in order to respond to product liability claims related to defects of the autonomous driving system, automotive manufacturers voluntarily comply with international standards for vehicle safety, such as ISO 26262 and ISO 21448, in the development process of autonomous vehicles. If an autonomous vehicle complying with these standards causes an accident, the automotive manufacturer could be freed from the product liability because it would be justified that the accident cannot be prevented even with state-of-the-art technologies which satisfy the safety standards. In the context of commercialization of autonomous vehicles, RSS and SFF, which are interpretable and mathematically rigorous, could be a candidate for the robust model substantiating that the safety of the intended functionality (SOTIF) of the planning subsystem is guaranteed in compliance with ISO 21448. Otherwise, without mathematical support \cite{Li 2020}, the reason for decision like “The trained network said so” cannot provide enough rational logic.

Intel released an open-source library called ad-rss-lib \cite{Gassmann 2019} that implements the RSS partially. Also, NVIDIA provides a software development kit called DriveWorks SDK that includes the SFF implementation \cite{Nister 2019} for approved users. Intel ad-rss-lib does not cover the whole scope of its paper, but it provides Python binding and CARLA \cite{Dosovitskiy 2017} integration \cite{Gassmann 2020}. However, NVIDIA DriveWorks SDK is a non-public property, and it is implemented to integrate with its NVIDIA DRIVE platform equipped with NVIDIA DRIVE OS \cite{NVIDIA 2019}, so researchers have utilized it relatively little. In the basic example architectures suggested by Intel and NVIDIA to integrate the RSS and SFF with existing autonomous driving systems, RSS and SFF play a role of the last resort to prevent collisions of the autonomous vehicle by overriding a decision from the planning subsystem.

Based on the NVIDIA’s conception, this work presents the SFF model implemented from scratch, integrating with open-source simulator CARLA. Furthermore, using the concept of claimed set and safety potential in the SFF implementation, we propose a method that integrates SFF into the planning subsystem to make a human-like driving policy that operates consistently whether safe or unsafe conditions regardless, eventually trying not to hinder the smooth traffic flow. Our work is different from NVIDIA’s example, which adds SFF, a separate ‘panic button’ module, into the existing system architectures to prevent collisions.



\section{Related Works}
\label{Section_2}
\textbf{Responsibility-Sensitive Safety (RSS)}, devised by Intel Mobileye in 2017 and released as an open-source library in 2019, is a mathematical model formalizing the safety abiding by 5 common sense rules among drivers. Basically, RSS is built on the time-to-collision (TTC) measure.

\textbf{Safety Force Field (SFF)}, devised by NVIDIA in 2017 and released only to permitted users as a software development kit in 2019, is also a mathematical model to guarantee the safety by trying to avoid unsafe situations. SFF is based on the measure for intersection of trajectories \cite{Alarcon 2019}.

As explained in Intel’s comparison table \cite{Intel 2019}, the concepts of Intel RSS and NVIDIA SFF are quite similar, and even example architectures presented by the two companies are identical in larger scheme. In the example, RSS or SFF is an upper-level add-on module aiming for vehicular safety and it can work regardless of established driving policy structure. Having compatibility, the module operates in parallel with planning subsystem, and acts as a restrictor that limits actions derived from the driving policy. See Appendix \ref{Appendix_A} for more information about RSS, SFF, and an example architecture.

However, in this architecture, the performance of the RSS or SFF module entirely depends on the completeness of the driving policy that developers implement. If the driving policy is not designed elaborately, the RSS or SFF will suddenly intervene as a contingency plan only when an accident is imminent, while driving normally with existing driving policy under safe conditions. This dichotomous system has an advantage of being relatively easy to design, but it makes itself being hard to expect consistent and smooth driving like a human driver in continuously changing circumstance.


\section{Claimed Set Predictor Learning and SFF Implementation}
\label{Section_3}
As described in Section \ref{Section_1}, by virtue of its interpretability and mathematical rigor, SFF can provide an obvious mathematical basis for maneuver decision of planning subsystem. The SFF module included in the NVIDIA DriveWorks SDK depends on the NVIDIA DRIVE OS, and the source code is not publicly disclosed. To utilize with CARLA simulator and TensorFlow platform, we implemented the SFF method from scratch. The details of SFF are described in NVIDIA’s whitepaper. In Appendix \ref{Appendix_B}, we explain the core concept and definitions of SFF in a nutshell.

As explained in Appendix \ref{Appendix_B}, calculating the safety potential is the core of SFF, but the safety potential is derived from the claimed set, and the claimed set is derived from safety procedure. Therefore, according to NVIDIA’s intention, the driving policy should be designed by developers in their own way first. Only after driving policy is completely implemented, SFF could be applied on autonomous driving system. In this context, to apply SFF, implementing the driving policy has the top priority.

Instead, in a different context, we tried to implement a driving policy using the safety potential of SFF. However, to obtain the claimed set required to calculate the safety potential, we fell into the contradiction that the safety procedure must already be implemented. To dispel the need to derive from a safety procedure, we propose a method learning the predictor for claimed sets. Taking advantage of CARLA simulator providing the ground truth of vehicle states, we trained the predictor network by supervised learning method. The predictor uses state vectors of vehicles and bird-eye view map image as inputs, and it outputs all vehicles’ 2D actions: x-axis acceleration and y-axis acceleration that make up the claimed set. By covering the claimed set with a smooth function like mollifier, we make the claimed set differentiable and as a result, get the safety potential by calculating the intersection area between the two actors’ claimed sets. In this way, without existing implemented safety procedure, we can obtain the claimed set of each actor vehicle and the safety potential. Our proposed system is described in Figure \ref{Figure_3_1}. See Appendix \ref{Appendix_C} for more claimed set prediction results.

\begin{figure}[h]
\centering
\includegraphics[width=0.67\linewidth]{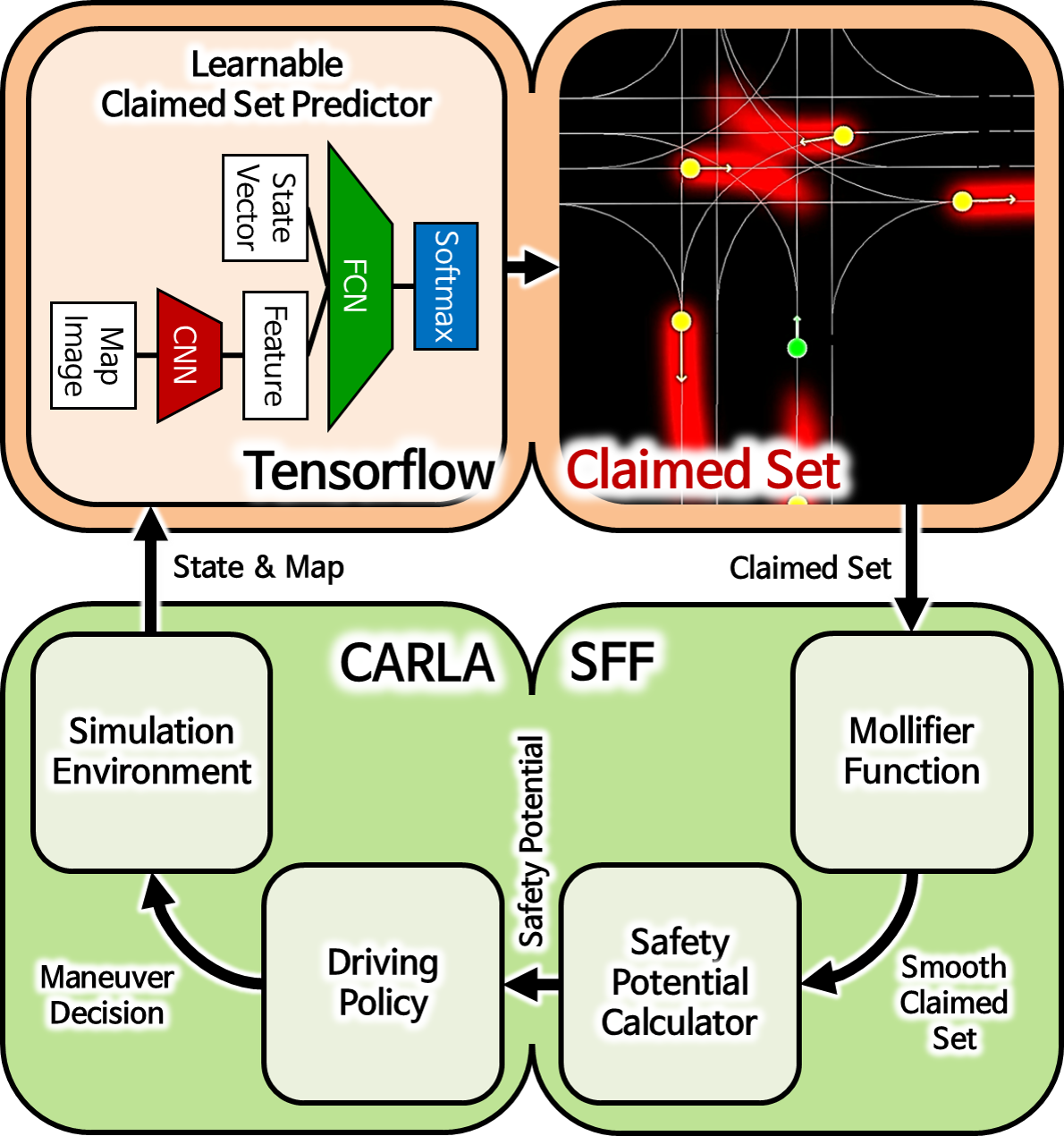} 
\caption{Simulation loop composed of proposed SFF-embedded CARLA. SFF integrates the claimed set predictor network for implementing the rationale-aware autonomous driving policy. (Red areas on upper-right image: other vehicles’ claimed sets predicted by trained predictor network.)}
\label{Figure_3_1}
\end{figure}


\section{SFF-Based Driving Policy Implementation}
\label{Section_4}
As described in Section \ref{Section_2}, with the dichotomous system based on NVIDIA’s basic example, the smoothness of maneuver comes from the completeness of driving policy. If the autonomous driving system has a crude driving policy, for instance, while cruising depending on the existing policy, even though there is a potential risk ahead, the SFF will intervene and take an abrupt action only at the moment when the risk is imminent, which is not ideal driving from human perspective \cite{Bae 2020}.

In Section \ref{Section_3}, we proposed the learning of SFF claimed set predictor represented in a deep neural network and the implementation of safety potential calculation. Based on our implementations, we can get claimed sets of all vehicles on CARLA simulation, and using those claimed sets, we can calculate the safety potential of the ego-vehicle. To verify that our proposed model is implemented and trained correctly, using the calculated safety potential, we can eventually make an integrated driving policy (safety procedure) that operates consistently without having two separate modules dealing with safe condition and unsafe condition respectively. Our integrated driving policy ultimately aims to be human-like, without being too dogmatic or passive to go with traffic flow smoothly \cite{Naumann 2019, Emuna 2020}.

To evaluate whether our SFF implementation is equivalent to existing models, we compare the performance of our driving policy with CARLA autopilot and RSS implementation on CARLA. We test an autonomous driving agent with our claimed set predictor to drive on road where other vehicles with random aggression are running. The random aggression means that sometimes other vehicles do not care about their surroundings and traffic signals according to probability hyperparameters. We divide the random aggression into 4 levels. In this experiment, we test how quickly and safely the agent arrives at destination without interfering with traffic flow on CARLA. In given period, we count the number of arrivals at randomly designated destinations and the accident-free timesteps of ego-vehicle. The results are represented in Table \ref{Table_4_1}. See Appendix \ref{Appendix_D} for experiment details.

\begin{table}[]
\caption{Experiment results. The number of arrivals represents efficiency, and the accident-free time represents safety. Each value is an average of 50 iterations. The higher the value, the better the performance of the corresponding model. (Bold: our result value.)}
\label{Table_4_1}
\centering
\renewcommand{\arraystretch}{1.16} 
\begin{tabular}{llll}
\toprule
Aggression                    & Driving Policy                  & Number of arrivals & Accident-free time \\ \hline
\multirow{4}{*}{No}           & No                              & 0.26               & 406                \\
                              & CARLA autopilot                 & 0.92               & 1690               \\
                              & RSS-CARLA implementation        & 0.30               & 2115               \\
                              & SFF-CARLA implementation \textbf{(Ours)} & \textbf{1.40}      & \textbf{2495}      \\ \hline
\multirow{4}{*}{Low}          & No                              & 0.18               & 333                \\
                              & CARLA autopilot                 & 0.86               & 1506               \\
                              & RSS-CARLA implementation        & 0.40               & 2427               \\
                              & SFF-CARLA implementation \textbf{(Ours)} & \textbf{1.36}      & \textbf{2287}      \\ \hline
\multirow{4}{*}{Intermediate} & No                              & 0.18               & 337                \\
                              & CARLA autopilot                 & 0.78               & 1436               \\
                              & RSS-CARLA implementation        & 0.26               & 2173               \\
                              & SFF-CARLA implementation \textbf{(Ours)} & \textbf{1.24}      & \textbf{2050}      \\ \hline
\multirow{4}{*}{High}         & No                              & 0.06               & 318                \\
                              & CARLA autopilot                 & 0.52               & 1332               \\
                              & RSS-CARLA implementation        & 0.44               & 2594               \\
                              & SFF-CARLA implementation \textbf{(Ours)} & \textbf{1.00}      & \textbf{1886}      \\ \bottomrule
\end{tabular}
\end{table}


\section{Discussion}
\label{Section_5}
In this work, we implement SFF, a mathematically proven model that could comply with ISO 21448 SOTIF, from scratch, and integrate it with CARLA simulator. Our method using the claimed set predictor does not need to care about an implementation method of existing driving policy to get claimed sets of vehicles. Using the claimed set predictor, we present an integrated driving policy that does not utilize an extra safety module. We verify that our driving policy based on predictor network shows competitiveness on safety compared with RSS by experiments at CARLA.

In future work, it might be possible to train not only the claimed set predictor but also the driving policy \cite{Zhao 2021} by reinforcement learning (RL) \cite{Nageshrao 2019, Baheri 2020, Cao 2022}, using the safety potential as a reward. The meta RL and the multi-agent RL \cite{Shalev-Shwartz_2016, Tang 2019, Zhu 2021} should also be considered to improve the driving policy for full automation. Furthermore, to specify the operational design domain (ODD) of autonomous driving system, it is essential to verify the system \cite{Tuncali 2019, Xu 2021} by detailed scenarios \cite{Shashua 2018, Ponn 2019, Vaskov 2019, Neurohr 2020, Calo 2020, Riedmaier 2020}, not randomly generated destination scripts.


\begin{ack}
This work was supported by the Institute for Information Communications Technology Planning Evaluation (IITP) grant funded by the Ministry of Science and ICT (MSIT, Korea, No.2021-0-01352, Development of technology for validating autonomous driving services in perspective of laws and regulations). And graduate student activities in this work were partly supported by Korea Institute for Advancement of Technology(KIAT) grant funded by the Korea Government(MOTIE)(P0020535, The Competency Development Program for Industry Specialist).

\end{ack}



\appendix

\section{RSS and SFF}
\label{Appendix_A}
\textbf{Responsibility-Sensitive Safety (RSS)} compares a dangerous time $t$ of the ego vehicle with the danger threshold time $t_{b}^{long}$, $t_{b}^{lat}$ on both longitudinal and lateral side. If the threshold is reached, RSS judges it as a dangerous situation and applies a proper response that follows the constraint on the speed with either a longitudinal or lateral acceleration. In other words, the threshold could be represented in a trajectory set polygon. If the trajectory sets between ego vehicle and other road user are intersected, RSS chooses one of the following three decisions to restore the safe condition: brake or continue forward or drive away.

\textbf{Safety Force Field (SFF)} says that if actors follow the safety procedure, which is a family of control policies, the safety potential $\rho_{AB}$ that quantifying the risk does not increase anymore, so it can be guaranteed that the actors will not cause unsafe situation eventually. This can be proved mathematically by a chain rule for the safety potential. In short, the method of RSS is to minimize the intersection between actors’ claimed sets which is an union of trajectories resulting from the each actor’s safety procedure. Key definitions are described in Appendix \ref{Appendix_B}.

\begin{figure}[h]
\includegraphics[width=\linewidth]{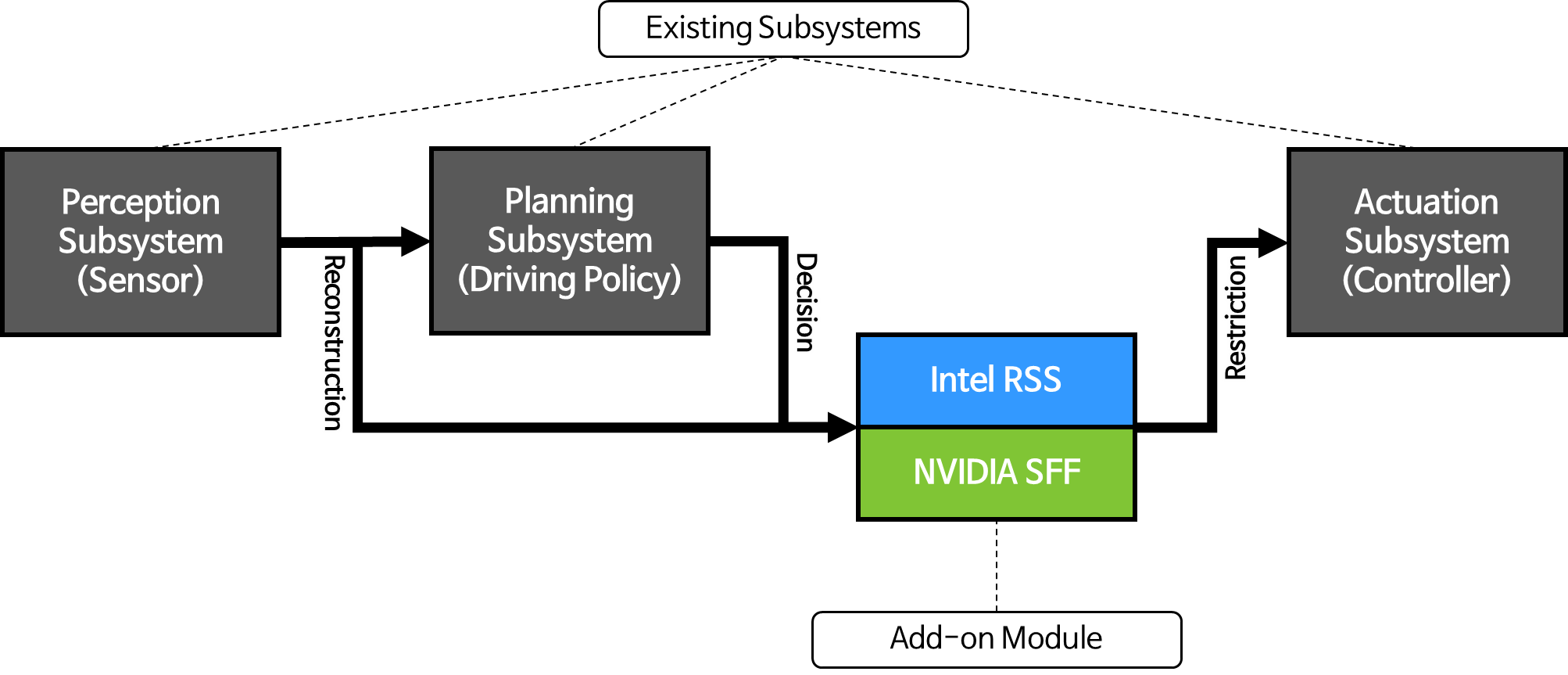}
\caption{Basic example architecture with RSS or SFF. (Grey: existing subsystems. Blue/Green: RSS/SFF implementation as an add-on module.)}
\label{Figure_A_1}
\end{figure}

Intel RSS or NVIDIA SFF is added as an add-on module harmonizing with existing subsystems. It receives both the world reconstruction data from perception subsystem and the maneuver decision from planning subsystem. For ego-vehicle's safety, as an upper-level restrictor, it could override received decision and pass a limited decision to actuation subsystem. A basic example architecture with RSS or SFF suggested by Intel and NVIDIA is described in Figure \ref{Figure_A_1}.

\section{Definitions in SFF}
\label{Appendix_B}
\textbf{State \boldmath$x_{A}(t)$} is a vector containing position (2D or 3D), direction, and velocity of the vehicle actor $A$.

\textbf{Control Policy \boldmath$\frac{dx_{A}}{dt}=f(x_{w},t)$} is a smooth and bounded function of differentiating the state $x_{A}(t)$ with respect to time $t$.

\textbf{Safety Procedure \boldmath$S_{A}=\{\frac{dx_{A}}{dt}\}$} is family of control policies $\frac{dx_{A}}{dt}$.

\textbf{Claimed Set \boldmath$C_{A}(x_{A})$} is a union of trajectories acquired by the safety procedure $S_{A}$. It is covered by a smooth function.

\textbf{Safety Potential \boldmath$\rho_{AB}$} is a non-negative measure of intersection between the claimed sets $C_{A}(x_{A})$ and $C_{B}(x_{B})$ of actor $A$ and $B$. It is a bump function where $\rho_{AB}=0$ if there is no intersection between claimed sets, and $\rho_{AB}>0$ if intersection occurs. The safety potential can be defined as a dot product between claimed sets, which are smooth functions, and the dot product between real number functions is computed as an integral. In simple terms, safety potential is an area where the claimed sets from road users intersect.

\textbf{Safety Potential \boldmath$F_{AB}=-\frac{d\rho_{AB}}{dx_{A}}$} is a negative gradient of safety potential $\rho_{AB}$. If actor $A$ and $B$ follow their respective safety procedures $S_{A}$ and $S_{B}$, the safety potential $\rho_{AB}$ does not increase anymore.


\section{Claimed Set Prediction Results}
\label{Appendix_C}
\begin{figure}[h]
\begin{subfigure}{0.5\textwidth}
\centering
\includegraphics[width=0.95\linewidth]{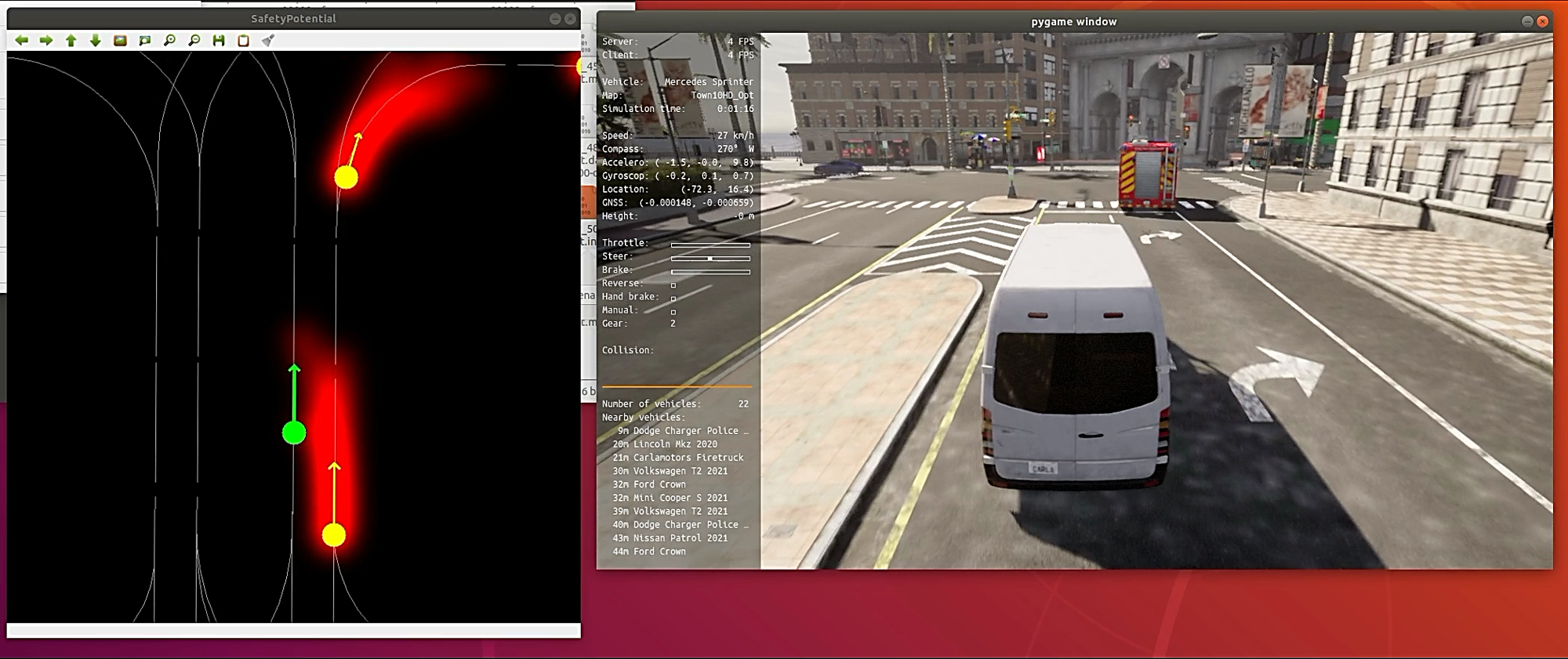}
\caption{}
\label{Figure_C_a}
\end{subfigure}
\begin{subfigure}{0.5\textwidth}
\centering
\includegraphics[width=0.95\linewidth]{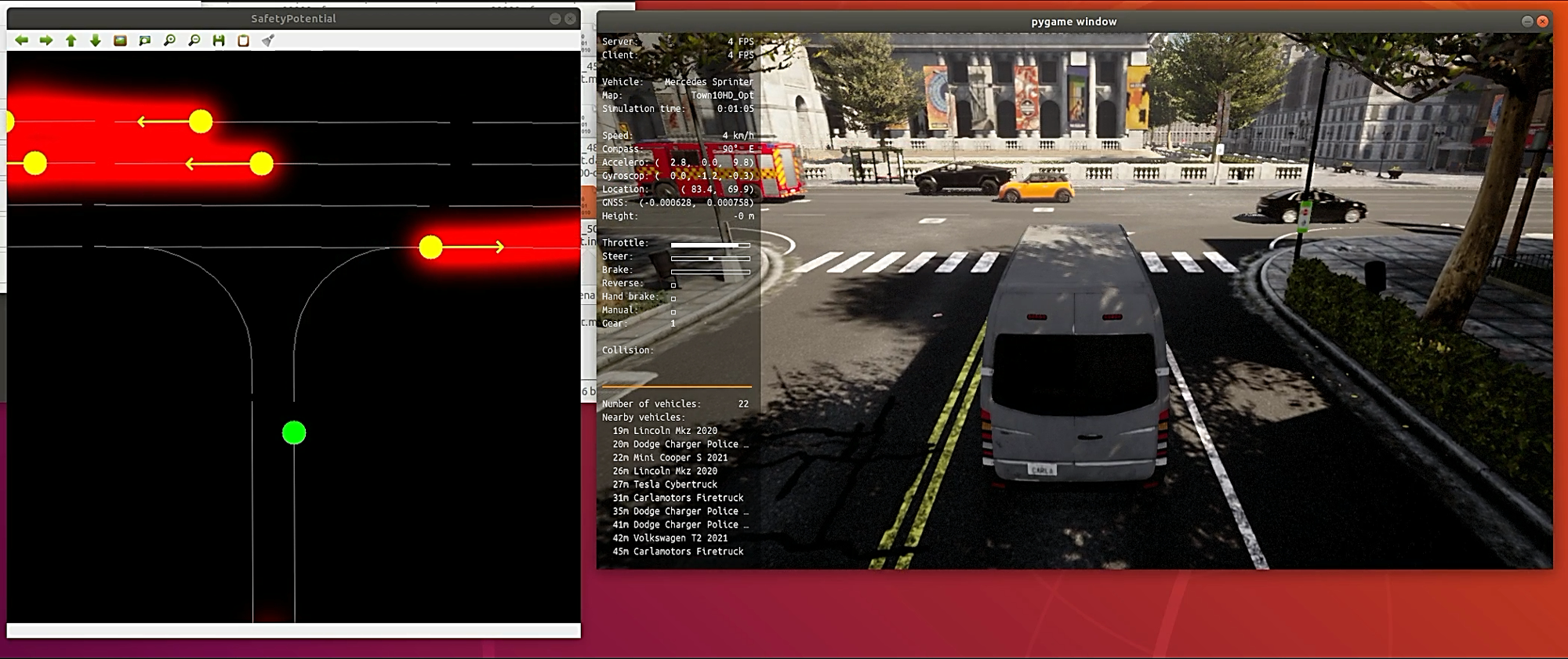}
\caption{}
\label{Figure_C_b}
\end{subfigure}
\begin{subfigure}{0.5\textwidth}
\centering
\includegraphics[width=0.95\linewidth]{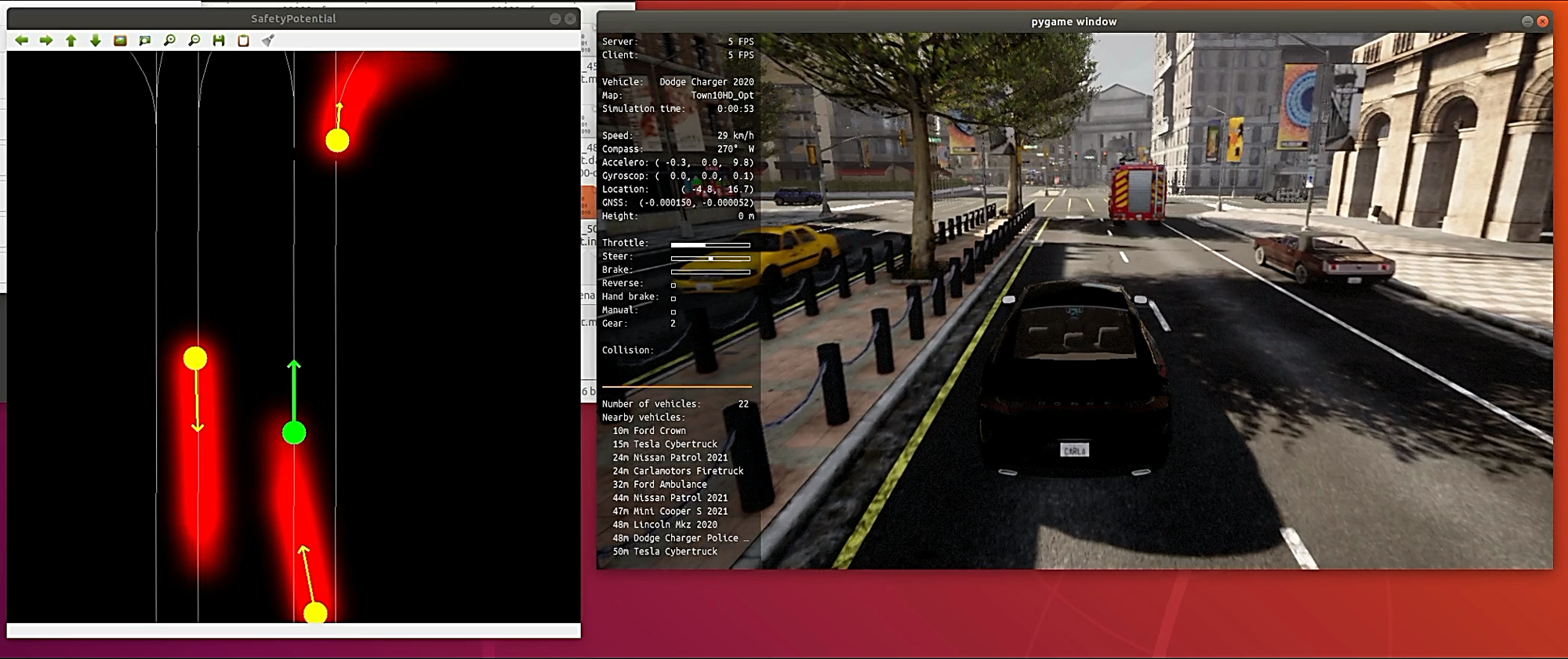}
\caption{}
\label{Figure_C_c}
\end{subfigure}
\begin{subfigure}{0.5\textwidth}
\centering
\includegraphics[width=0.95\linewidth]{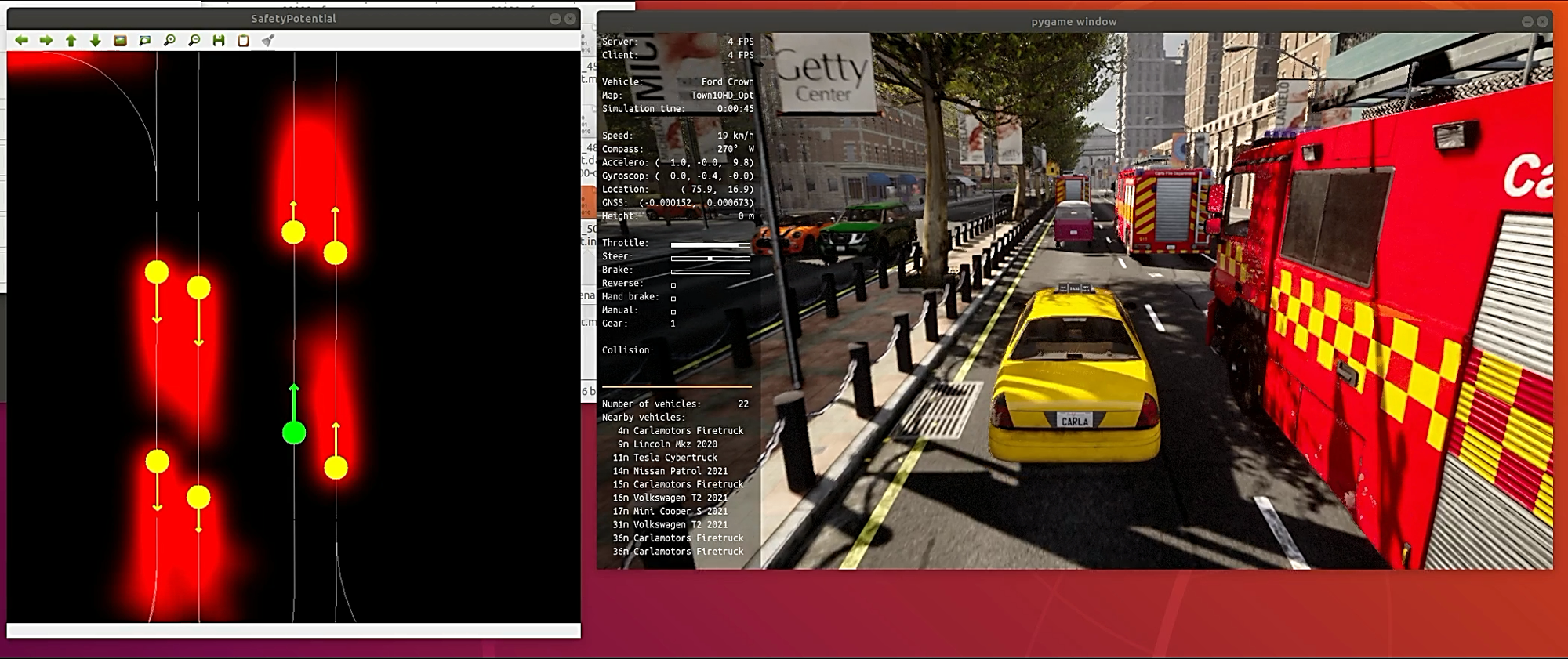}
\caption{}
\label{Figure_C_d}
\end{subfigure}
\caption{Visualization of claimed set prediction and corresponding scene of CARLA simulator. (Green dot: ego-vehicle's position. Yellow dots: other vehicles' positions. Red areas: other vehicles' claimed sets predicted by trained predictor network and smoothed by mollifier function.)}
\label{Figure_C_1}
\end{figure}


\section{Experiment Details}
\label{Appendix_D}
\paragraph{Random Aggression}
The aggression level is divided into 4 levels: No, Low, Intermediate, and High. Each level has different probability hyperparameters deciding the disregard level for two situations: surroundings during lane changing, and traffic signals during crossroad passing. The “No” level has the lowest probability, 0 exactly, and the “High” level has the highest.

\begin{figure}[h]
\centering
\includegraphics[width=0.8\linewidth]{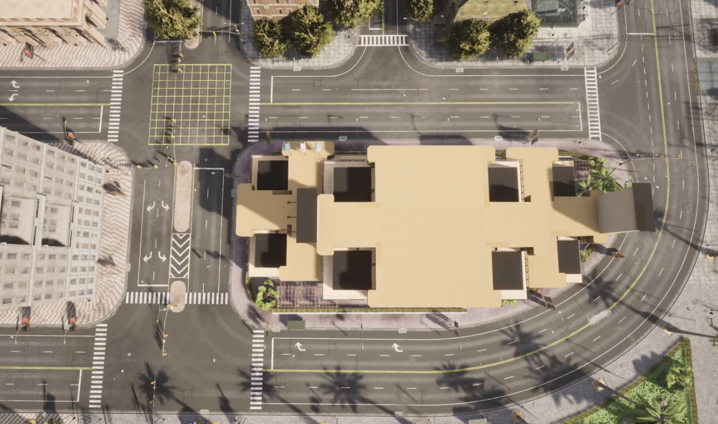}
\caption{CARLA Town10 environment used for training and test in this work.}
\label{Figure_D_1}
\end{figure}

\paragraph{Environment}
As an environment, we use the Town10 map provided by CARLA, a small town with 9 crossroads described in Figure \ref{Figure_D_1}. There are 50 other vehicles as non-player characters (NPC) on the road. The vehicle models include sedan, SUV, van, and fire engine, which have different dimensions and dynamics. Each vehicle drives to a random destination. While driving, they decide to change their lane with a fixed probability, and while lane changing, sometimes they ignore their surroundings according to the aggression level hyperparameter. They also ignore traffic lights occasionally according to the aggression level.

\paragraph{Iteration}
We run 50 iterations for each random aggression level environment for each decision model. Each iteration consists of 5000 timesteps, but if the ego-vehicle is involved in a collision, the iteration terminates immediately. Every iteration is initialized randomly.

\paragraph{Driving Policy}
We implement a driving policy using the safety potential, which depends on the learnable claimed set predictor. The longitudinal controller decides the acceleration based on the safety potential of ego-vehicle, and the lateral controller decides the steering angle trying to follow an imaginary center line.

\end{document}